%% file: main.tex
\documentclass[10pt,twocolumn,letterpaper]{article}

\usepackage{iccv}
\usepackage{times}
\usepackage{epsfig}
\usepackage{graphicx}
\usepackage{amsmath}
\usepackage{amssymb}

\usepackage{graphicx}
\usepackage{amsmath}
\usepackage{amssymb}
\usepackage{booktabs}
\usepackage{bbm}
\usepackage{graphicx}
\usepackage{cuted}
\usepackage{subcaption}

\usepackage[pagebackref=true,breaklinks=true,letterpaper=true,colorlinks,bookmarks=false]{hyperref}

\usepackage{xcolor}

\usepackage[capitalize]{cleveref}
\crefname{section}{Sec.}{Secs.}
\Crefname{section}{Section}{Sections}
\Crefname{table}{Table}{Tables}
\crefname{table}{Tab.}{Tabs.}

\iccvfinalcopy 

\def\iccvPaperID{5808} 

\ificcvfinal\fi

\begin{document}

\title{Parametric Depth Based Feature Representation Learning for Object Detection and Segmentation in Bird’s-Eye View}

\author{Jiayu Yang$^{1,3^*}$,\;\; Enze Xie$^2$,\;\; Miaomiao Liu$^1$,\;\; Jose M. Alvarez$^3$\\
$^1$Australian National University, $^2$The University of Hong Kong, $^3$NVIDIA\\
{\tt\small \{jiayu.yang, miaomiao.liu\}@anu.edu.au,}\;\;{\tt\small xieenze@connect.hku.hk,}\;\;{\tt\small josea@nvidia.com}
}

\maketitle
\ificcvfinal\thispagestyle{empty}\fi

\begin{strip}
  \centering
  \vspace{-1.6cm}
\includegraphics[width=1.0\linewidth]{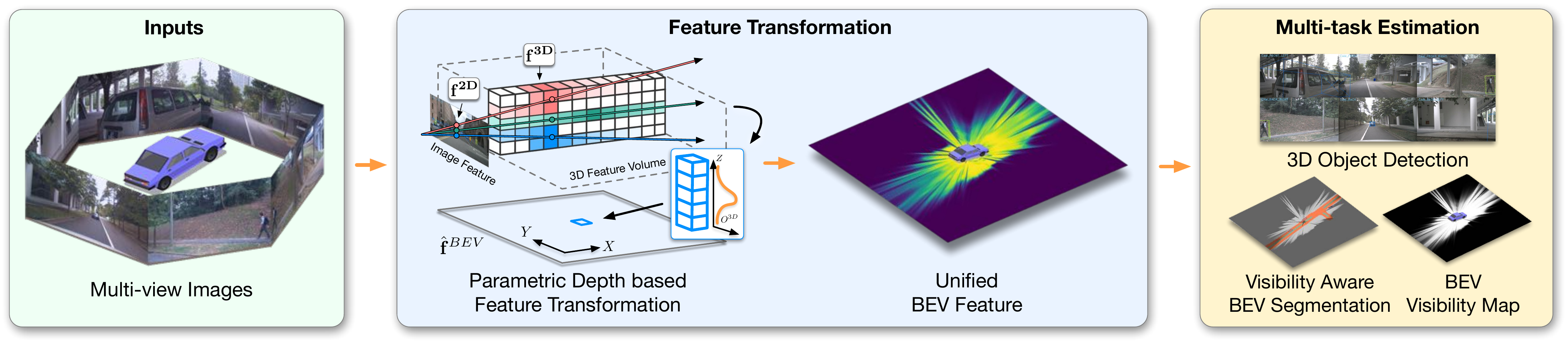}\\
  \vspace{-0.2cm}
  \captionof{figure}{Given multi-view images and camera parameters, our framework utilize parametric depth to transform image feature into BEV space for jointly estimating 3D object detection, BEV segmentation and a BEV visibility map.}
  \vspace{-0.3cm}
  \label{fig:teaser}
\end{strip}

\input{abstract}

\input{introduction}
\input{relatedwork}
\input{method}
\input{experiments}
\input{conclusion}

{\small
\bibliographystyle{ieee_fullname}
\bibliography{egbib}
}

\end{document}

%% file: abstract.tex
\begin{abstract}
\vspace{-0.3cm}
Recent vision-only perception models for autonomous driving achieved promising results by encoding multi-view image features into Bird's-Eye-View (BEV) space. A critical step and the main bottleneck of these methods is transforming image features into the BEV coordinate frame. This paper focuses on leveraging geometry information, such as depth, to model such feature transformation. Existing works rely on non-parametric depth distribution modeling leading to significant memory consumption, or ignore the geometry information to address this problem. In contrast, we propose to use parametric depth distribution modeling for feature transformation. We first lift the 2D image features to the 3D space defined for the ego vehicle via a predicted parametric depth distribution for each pixel in each view. Then, we aggregate the 3D feature volume based on the 3D space occupancy derived from depth to the BEV frame. Finally, we use the transformed features for downstream tasks such as object detection and semantic segmentation. Existing semantic segmentation methods do also suffer from an hallucination problem as they do not take visibility information into account. This hallucination can be particularly problematic for subsequent modules such as control and planning. To mitigate the issue, our method provides depth uncertainty and reliable visibility-aware estimations.
\let\thefootnote\relax\footnote{$^*$The work is done during an internship at NVIDIA}
We further leverage our parametric depth modeling to present a novel visibility-aware evaluation metric that, when taken into account, can mitigate the hallucination problem. 
Extensive experiments on object detection and semantic segmentation on the nuScenes datasets demonstrate that our method outperforms existing methods on both tasks. 
\end{abstract}

%% file: introduction.tex
\section{Introduction}

In autonomous driving, multiple input sensors are often available, each of which has its coordinate frame, such as the coordinate image
frame used by RGB cameras or the egocentric coordinate frame used by the Lidar scanner. Downstream tasks, such as motion planning, usually require inputs in a unified egocentric coordinate system, like the widely used Bird's Eye View (BEV) space. Thus, transforming features from multiple sensors into the BEV space has become a critical step for autonomous driving. Here, we focus on this transformation for the vision-only setup where we take as input multi-view RGB images captured in a single time stamp by cameras mounted on the ego vehicle and output estimation results, such as object detection and segmentation, in a unified BEV space, see Fig.~\ref{fig:teaser}. 
In general, accurate depth information is crucial to achieve effective transformations.

\begin{figure}[t]
	\begin{center}
    \includegraphics[width=0.9\linewidth]{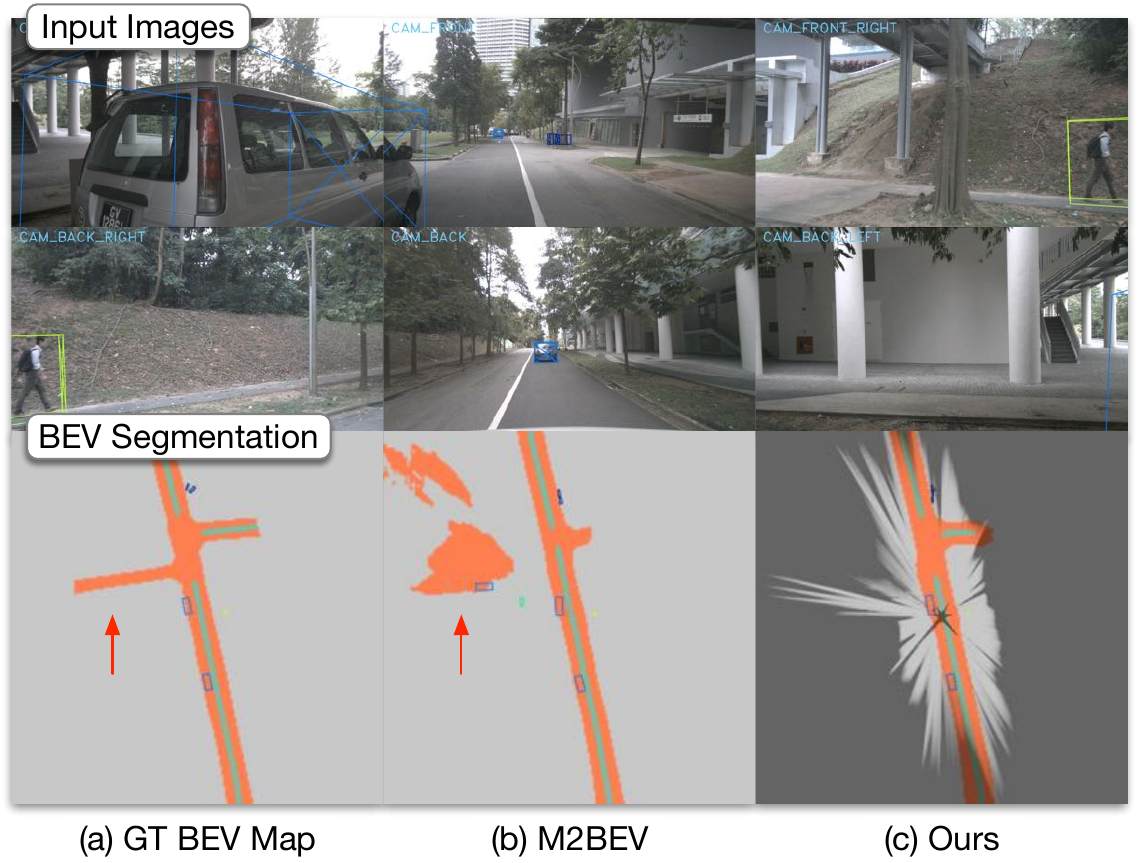}
    \vspace{-0.2cm}
	\caption{Hallucination in semantic segmentation. Current methods use ground truth obtained from maps (a) and therefore, predicted outputs (b) might represent parts that are not visible to the camera. As information is actually not available, it is not possible to determine if the road areas in the occluded areas is actually free for driving. Our approach enables creating a Visibility map (c) to decouple areas that are totally occluded to the camera from those that are actually visible.}
	\label{fig:hallucination}
	\end{center}
	\vspace{-0.9cm}
\end{figure}

Early methods\cite{pan2020vpn,pon} forgo explicit depth estimation and learn implicit feature transformations using neural networks, which suffers from the generalization problem since the neural network does not have an explicit prior of the underlying geometric relations. More recent methods \cite{lss, m2bev} adopt explicit but simplified depth representations for the transformation, which either requires large memory consumption, limiting the resolution~\cite{lss}; or over-simplifies the representation leading to noise in the BEV space\cite{m2bev}. Moreover, these simplified depth representation do not have the ability to efficiently provide visibility information. As downstream tasks such as semantic segmentation are trained using aerial map ground truth, the lack of visibility estimation usually results in hallucination effects where the network segments areas that are not visible to the sensor~\cite{lss, m2bev}, see Figure~\ref{fig:hallucination}. As a consequence, {those estimations can mislead downstream planning tasks as it is extremely dangerous to drive towards hallucinated road but actually non-driveable, especially in high speed.}

To address these limitations, we propose to adopt explicit parametric depth representation and geometric derivations as guidance to build a novel
feature transformation pipeline. We estimate a parametric depth distribution and use it to derive both a depth likelihood map and an occupancy distribution to guide the transformation from image features into the BEV space. Our approach consists of two sequential modules: a geometry-aware feature lifting module and an occupancy-aware feature aggregation module. Moreover, our parametric depth-based representation enables us to efficiently derive a visibility map in BEV space, which provides valuable information to decouple visible and occluded areas in the estimations and thus, mitigate the hallucination problem. We also derive ground-truth visibility in BEV space, which enables us to design a novel evaluation metric for BEV segmentation that takes visibility into account and reveals insight of selected recent methods~\cite{lss, m2bev} in terms of estimation on visible region and hallucination on occluded region.

Our contributions can be summarized as follows:
\begin{itemize}
    \item We propose a geometry-aware feature transformation based on parametric depth distribution modeling to map multi-view image features into the BEV space. Our depth distribution modeling enables the estimation of visibility maps to decouple visible and occluded areas for downstream tasks.
    \item The proposed feature transformation framework consists of a novel feature lifting module that leverages the computed depth likelihood to lift 2D image features to the 3D space; and a feature aggregation module to project feature to the BEV frame through the derived 3D occupancy.
    \item We further propose a novel visibility-aware evaluation metric for segmentation in BEV space that reveals the insight of estimation on visible space and hallucination on occluded space. 
\end{itemize}
Extensive experiments on the nuScenes dataset on object detection and semantic segmentation demonstrate the effectiveness of our method yielding state of the art results for these two tasks with a negligible compute overhead. 

%% file: relatedwork.tex
 \begin{figure*}[!t]
	\begin{center}
     \vspace{-0.3cm}
    \includegraphics[width=1.0\linewidth]{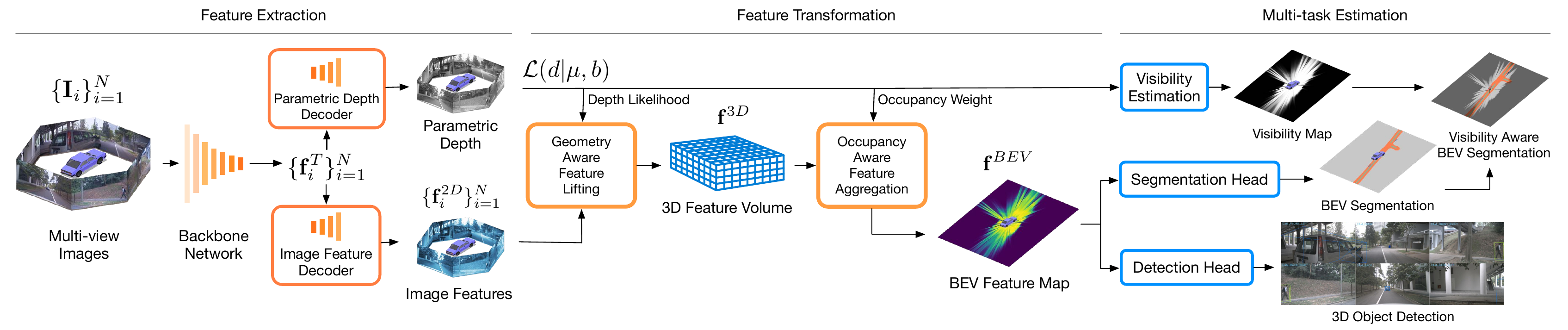}
    \vspace{-0.5cm}
	\caption{\textbf{Proposed framework}. The main novelties are a parametric depth decoder in the feature extraction, a geometry-aware feature lifting module and an occupancy aware feature aggregation in the feature transformation. We also introduce a visibility estimation module as part of the multi-task estimation.}
	\label{fig:pipeline}
	\end{center}
	\vspace{-0.7cm}
\end{figure*}

\section{Related Work}
\noindent\textbf{External depth based feature transformations.}
When given depth input either from Lidar sensor or stereo matching, image feature can easily be transformed into BEV space\cite{lang2019pointpillars, shi2019pointrcnn, shi2020pv, yan2018second, yin2021center}. PointPillar\cite{lang2019pointpillars} extract features from a 3D point cloud and aggregate the features into BEV space. PseudoLidar\cite{wang2019pseudo, qian2020end} based methods firstly estimate a depth using stereo matching given stereo image pair as input followed by unprojecting the feature based on estimated depth. However, in real-life applications, Lidar sensors or stereo image inputs are not always available, which limits these line of methods.

\noindent\textbf{Feature transformations without reliable depth input.}
 Without reliable depth input, various feature transformation methods have been proposed\cite{chen2016monocular, kehl2017ssd, poirson2016fast, qin2019monogrnet, simonelli2019disentangling}, starting from early methods\cite{pan2020vpn, pon} that learn implicit feature transformations using neural networks. Learned transformation can suffer from the generalization problem, since the neural network does not explicitly account for changes in cameras' intrinsic and extrinsic parameters. Recent methods \cite{lss, m2bev, wang2022detr3d} adopt various depth representations to explicitly transform features based on multi-view geometry to the BEV space. The key in these methods is the underlying depth representation, which dominates the resolution and accuracy the feature transformation module can achieve. For instance, LSS~\cite{lss} adopts a non-parametric depth representation. It represents depth as a discretized probability density function along each visual ray, which can be treated as a categorical distribution of depth. It can further form the depth probability volume in LSS for all pixels in an image. When the sampling rate is sufficient, such non-parametric depth distribution can adequately represent a large variety of depths, including multi-modal depth distributions. In practice, however, to estimate such depth representation, the backbone needs to estimate a probability volume that is cubic with the input image size and increases significantly along the number of input images, which limits the image and depth resolution.

To address this limitation, M$^2$BEV~\cite{m2bev} adopts a simplified depth representation assuming the depth of all pixels follows a uniform distribution. Under this assumption, features are directly lifted to every location on the visual ray, resulting identical feature along the entire ray with no difference. Following works \cite{li2022bevformer,bartoccioni2022lara} followed similar depth representation. Such simplified representation have advantage on efficiency, as the backbone network do not need to estimate any parameter for the depth, but can cause ambiguity and noise in the 3D space. 

Unlike the non-parametric depth distribution used in \cite{lss} or the uniform depth distribution in M2BEV\cite{m2bev}, we adopt a parametric depth distribution to model pixel-wise depth for feature lifting. Parametric depth distribution represents depth as a continuous distribution such as Gaussian or the Laplacian distribution, and its estimated distribution parameters can be used to evaluate depth likelihood or depth probability on any given depth value along each ray. To model the depth for a pixel, it takes only two parameters ($\mu,\sigma$) for Gaussian and two ($\mu,b$) for Laplacian, so it can be more efficient than non-parametric distribution. Moreover, its continuous nature allows evaluating depth likelihood on any points along the visual ray, which can achieve a higher depth resolution than the diescretized non-parametric distribution. We specifically designed our pipeline incorporating parametric depth to improve 2D-BEV feature transformation and also propose the derivation of visibility for subsequent planning tasks and visibility-aware evaluations.

 \noindent\textbf{Aggregating 3D feature into BEV space.} Given the lifted feature in 3D space, most existing works including LSS~\cite{lss} and M$^2$BEV~\cite{m2bev} use the feature concatenation method introduced by Pointpillars\cite{lang2019pointpillars} for transforming 3D features into BEV space. The 3D feature volume is split along horizontal dimensions and interpreted as pillars of features. Then, a feature vector is created by concatenating features along the vertical dimension for each pillar. All the concatenated features form a 2D feature map, which is converted into BEV feature map by few convolution layers. This design allows each voxel along the Z-axis to have equal contribution to the final BEV feature. However, this method can be affected by noisy features on empty spaces. We thus propose to compress the features based on a calculated space occupancy probability from the parametric depth distribution. Our proposed method can largely reduce the influence of those empty voxels to the aggregated features.

\noindent\textbf{Joint Detection and Segmentation in BEV space.}
M$^2$BEV recently proposed a unified detection and segmentation framework in BEV space, which we leverage to evaluate the effectiveness of our method. Specifically, the image features are transformed into a unified BEV feature, which is used by two parallel heads, a detection head and a segmentation head, to achieve multi-task estimation. M$^2$BEV leverage a detection head design from Lidar-based detection methods \cite{lang2019pointpillars} and modify it to better suit camera-based methods. Their segmentation head is inspired by the design from \cite{lss}. However, in contrast to prior work, we leverage the proposed explicit feature transformations based on parametric depth to address its weaknesses. 

\noindent\textbf{Temporal extension.}
Few concurrent methods \cite{li2022bevformer, liu2022petrv2, zhang2022beverse, huang2022bevdet4d, wang2022monocular,park2022time,hu2022st} proposed to utilize temporal information to further boost segmentation and detection performance in BEV space and achieved promising results. Most of these methods, including BEVFormer\cite{li2022bevformer}, BEVerse\cite{zhang2022beverse}, BEVDet4D\cite{huang2022bevdet4d} are based on the feature transformation module in LSS\cite{lss}. 
\cite{li2022bevdepth,li2022bevstereo} adopt depth supervision and temporal stereo matching to improve depth quality and further propose a more efficient implementation of LSS's Lift-splat step. \cite{liu2022petrv2,li2022bevformer,bartoccioni2022lara} query 2D features from projected location of 3D voxels, which does not explicitly use depth and is similar to the uniform depth assumption in M$^2$BEV\cite{m2bev}. Our contributions focusing on depth representation, feature transformation and visibility estimation is orthogonal to the temporal extension of these methods and our method can potentially be applied to these methods to further boost their performance and enable the efficient visibility inference.

%% file: method.tex

\section{Method}
Let us now introduce our framework to jointly perform segmentation and object detection. Shown in Fig.~\ref{fig:pipeline}, our framework comprised of three fundamental components: feature extraction, feature transformation, and multi-task estimation. The framework's key contributions include a parametric depth decoder integrated into the feature extraction, a geometry-aware feature lifting module, and an occupancy-aware feature aggregation module. Furthermore, we introduce a visibility estimation module as a constituent of the multi-task estimation that provide crucial visibility information for down-streaming planning tasks.

\subsection{Problem Statement}
Let~$\{{\bf I}_i\} _{i=1}^N,~{\bf I}_i\in\mathbb{R}^{H\times W \times 3}$,
be a set of RGB images taken at the same time slot, $H$ and $W$ define the image dimension, and $\{{\bf K}_i, {\bf R}_i, {\bf T}_i\}_{i=1}^N$ represent the intrinsic and extrinsic parameters for their corresponding camera poses, respectively. We focus on lifting the image features ${\bf f}_i^{2D}\in \mathbb{R}^{H\times W \times CH}$ to the 3D space as ${\bf f}^{\text{3D}}\in \mathbb{R}^{X'\times Y' \times Z'\times CH}$ and then aggregate them to the BEV space as ${\bf f}^{\text{BEV}}\in \mathbb{R}^{X\times Y \times CH_{B} }$ for 3D object detection and segmentation. 

\begin{figure}[!t]
	\begin{center}
\includegraphics[width=0.8\linewidth]{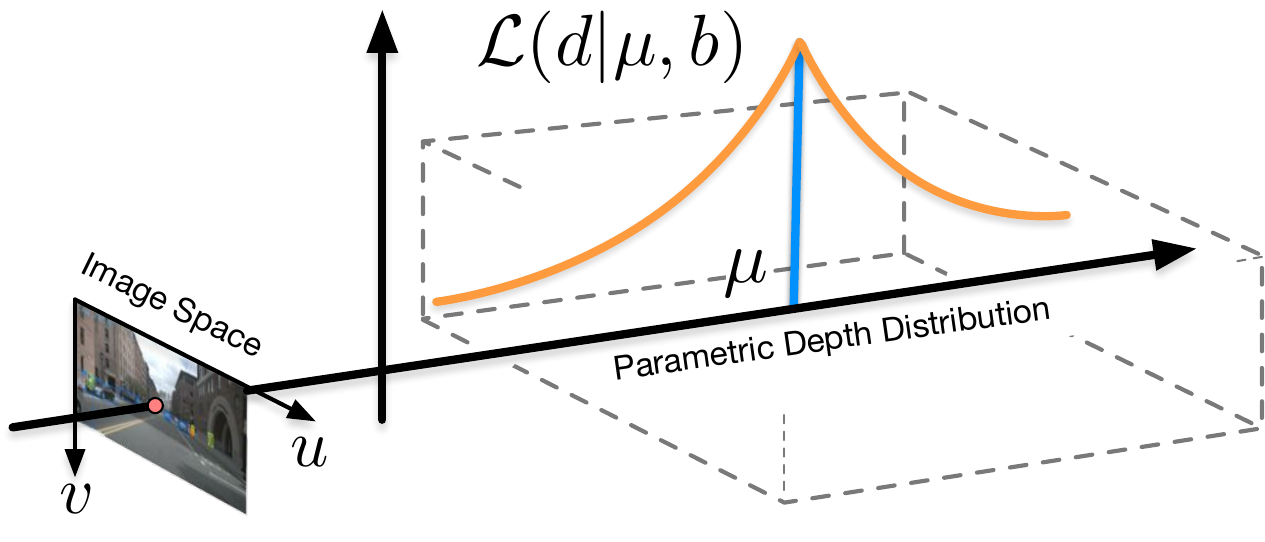}
	\caption{\textbf{Parametric depth distribution modeling.} We model depth using a Laplacian distribution. }
	\label{fig:pdepth}
	\end{center}
	\vspace{-0.4cm}
\end{figure}

\begin{figure}[!t]
	\begin{center}
    \includegraphics[width=0.8\linewidth]{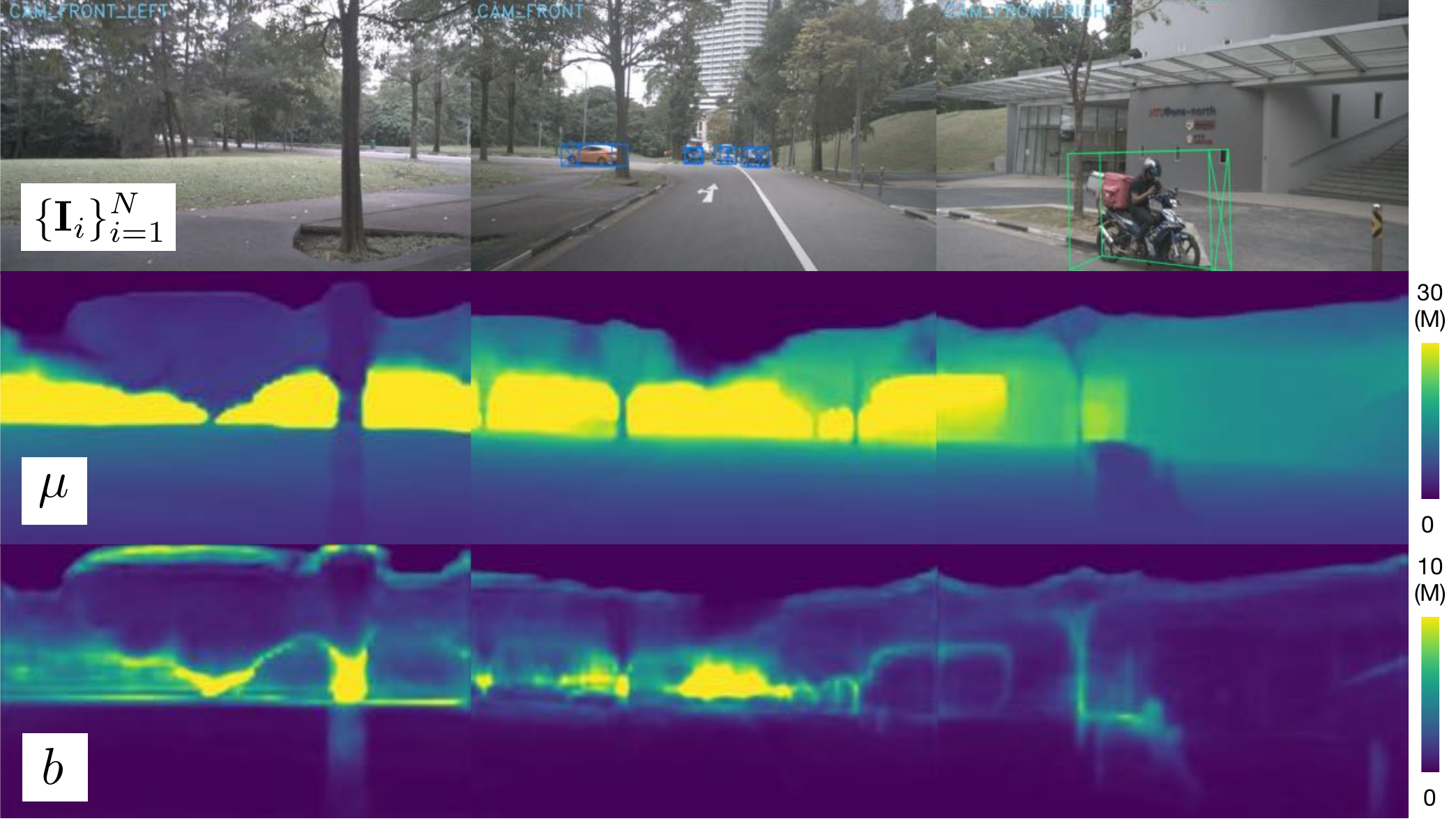}
	\caption{\textbf{Example of estimated parametric depth-distribution.} From top to bottom: input image, the estimated depth ($\mu$) and the diversity parameter ($b$) interpreted as the uncertainty of the estimation.}
	\label{fig:vis_pdepth}
	\end{center}
	\vspace{-0.6cm}
\end{figure}

\subsection{Parametric Depth Distribution Modelling}\label{sec:depth}
Let us first introduce our parametric depth distribution modelling. Given an image ${\bf I}_i$, we extract its latent features ${\bf f}_i^{T}$ using a backbone network followed by a image feature decoder network to extract 2D image features, ${\bf f}_i^{2D}$, see fig.~\ref{fig:pipeline}. Then, following depth estimation methods \cite{tosi2021smd,chen2019over}, we adopt a Laplacian distribution to model depth in real-world scenarios where the depth distribution for each pixel is given by,
\begin{equation}
    \mathcal{L}(d|\mu,b) = \frac{1}{2b}\exp(-\frac{|d-\mu|}{b}),
\end{equation}
\noindent where $\mu$ provides an estimation of the depth, and $b$ is the diversity parameter of the distribution, see Fig.~\ref{fig:pdepth}. The goal in this module is to estimate $(\mu, b)$.

 We design the parametric depth decoder network $\Phi_\theta$ to map the latent feature to the parameter space of the depth distribution: $\Phi_\theta: \mathbb{R}^{H\times W\times CH_{T}}\rightarrow \mathbb{R}^{H\times W\times 2},$
\noindent where $CH_T$ is the latent feature dimension. Note that when the ground-truth depth for each pixel is known, the depth distribution becomes a delta function, where the depth probability $p(d_{gt})$ on ground-truth depth $d_{gt}$ is one and zero anywhere else. However, in practice, the depth is unknown for each pixel. Given our modelled depth distribution, we can calculate the depth likelihood analytically based on our parametric modelling.
Fig.~\ref{fig:vis_pdepth} shows an example of depth distribution where $\mu$ gives an estimate of the depth and $b$ could be interpreted as the uncertainty of each estimation. Larger values of $b$ correspond to areas where the estimation is more uncertain.

\subsection{Geometry-aware Feature Lifting}\label{sec:lifting}
\begin{figure}[!t]
	\begin{center}
\includegraphics[width=0.7\linewidth]{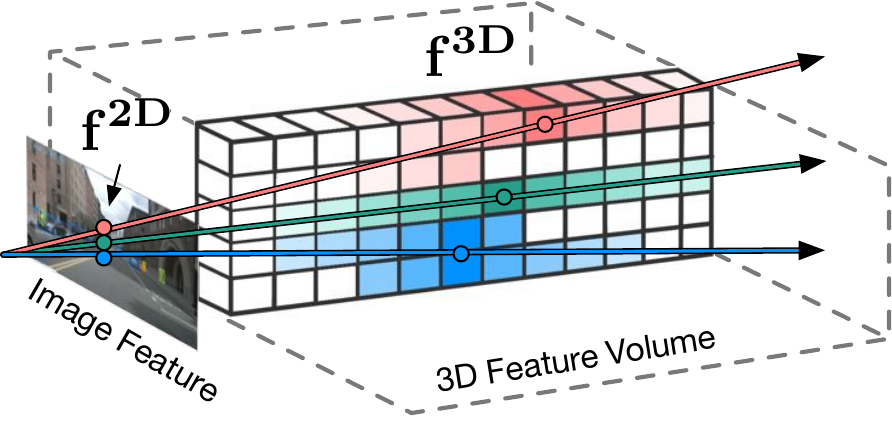}
	\caption{Geometry-aware feature lifting}
	\label{fig:lifting}
	\end{center}
	\vspace{-0.9cm}
\end{figure}

Fig.~\ref{fig:lifting} depicts our geometry-aware feature lifting module to transform the 2D image features ${\bf f}_i^{2D} \in \mathbb{R}^{H\times W\times CH}$ from the camera coordinate system into 3D space defined for the ego vehicle coordinate system, generating the 3D feature volume ${\bf f}_i^{3D} \in \mathbb{R}^{X'\times Y'\times Z'\times CH_{I}}$.  

Ideally, the 2D image feature for each pixel is back-projected along the visual ray to the 3D location defined by its ground truth depth value $ {\bf f}^{3D}({\bf P}_{gt}) = {\bf f}^{2D}({\bf p})$, where ${\bf P}_{gt} = d_{gt}{\bf K}_i^{-1}\tilde{\bf p}$, $\tilde{\bf p}$ is the homogeneous coordinate for ${\bf p}$. Without knowing the true depth value for each pixel, we discretize the 3D space into voxels and thus aggregate the feature for each voxel by forward projecting it to multi-view images. 

Precisely, let ${\bf P}_j = (x_j, y_j, z_j)^{T}$ define the 3D coordinate of centre for voxel $j$. Given the camera poses for multiple views, we project it to image ${\bf I}_i$ as 
$d^i_j\tilde{{\bf p}}^i_j = {\bf K}_i({\bf R}_i\tilde{{\bf P}}_j+{\bf T}_i)$ where $\tilde{{\bf p}}^i_j$ denotes the homogenous coordinate of ${\bf p}^i_j$ in image ${\bf I}_i$. Meanwhile, we can obtain the depth value of ${\bf P}_j$ in view $i$ as $d^i_j$. Based on our parametric depth modelling, we obtain the likelihood of $d^i_j$ being on the object surface as 
\begin{equation}\label{eq:likelihood}
    \alpha_{d^i_j} = \mathcal{L}(d^i_j|\mu^i_{{\bf p}^i_j},b^i_{{\bf p}^i_j}) = \frac{1}{2b^i_{{\bf p}^i_j}}\exp(-\frac{|d^i_j-\mu^i_{{\bf p}^i_j}|}{b^i_{{\bf p}^i_j}}).
\end{equation}
We similarly project the voxel to all views and aggregate the feature for the $j$-th voxel as 

\begin{equation}
    {\bf f}_j^{3D} = \sum_{i=1}^N\alpha_{d^i_j} {\bf f}_i^{2D}({\bf p}^i_j),
\end{equation}
where ${\bf f}_i^{2D}$ is the extracted image feature. We adopts bilinear interpolation to obtain ${\bf f}_i^{2D}({\bf p}^i_j)$ when ${\bf p}^i_j$ is a non-grid coordinate. All lifted 3D features form the 3D feature volume ${\bf f}^{3D} \in \mathbb{R}^{X'\times Y'\times Z'\times CH}$, which is then aggregated by our occupancy aware feature aggregation module into 2D BEV feature, introduced in the following section.

\subsection{Occupancy-aware Feature Aggregation}\label{sec:compression}

Our occupancy-aware feature aggregation module aggregates the 3D feature volume ${\bf f}^{3D} \in \mathbb{R}^{X'\times Y'\times Z'\times CH}$ from ego vehicle 3D coordinate frame into BEV space, forming BEV feature map ${\bf f}^{BEV} \in \mathbb{R}^{X\times Y\times CH_B}$.

\begin{figure}[!t]
	\begin{center}
\includegraphics[width=0.8\linewidth]{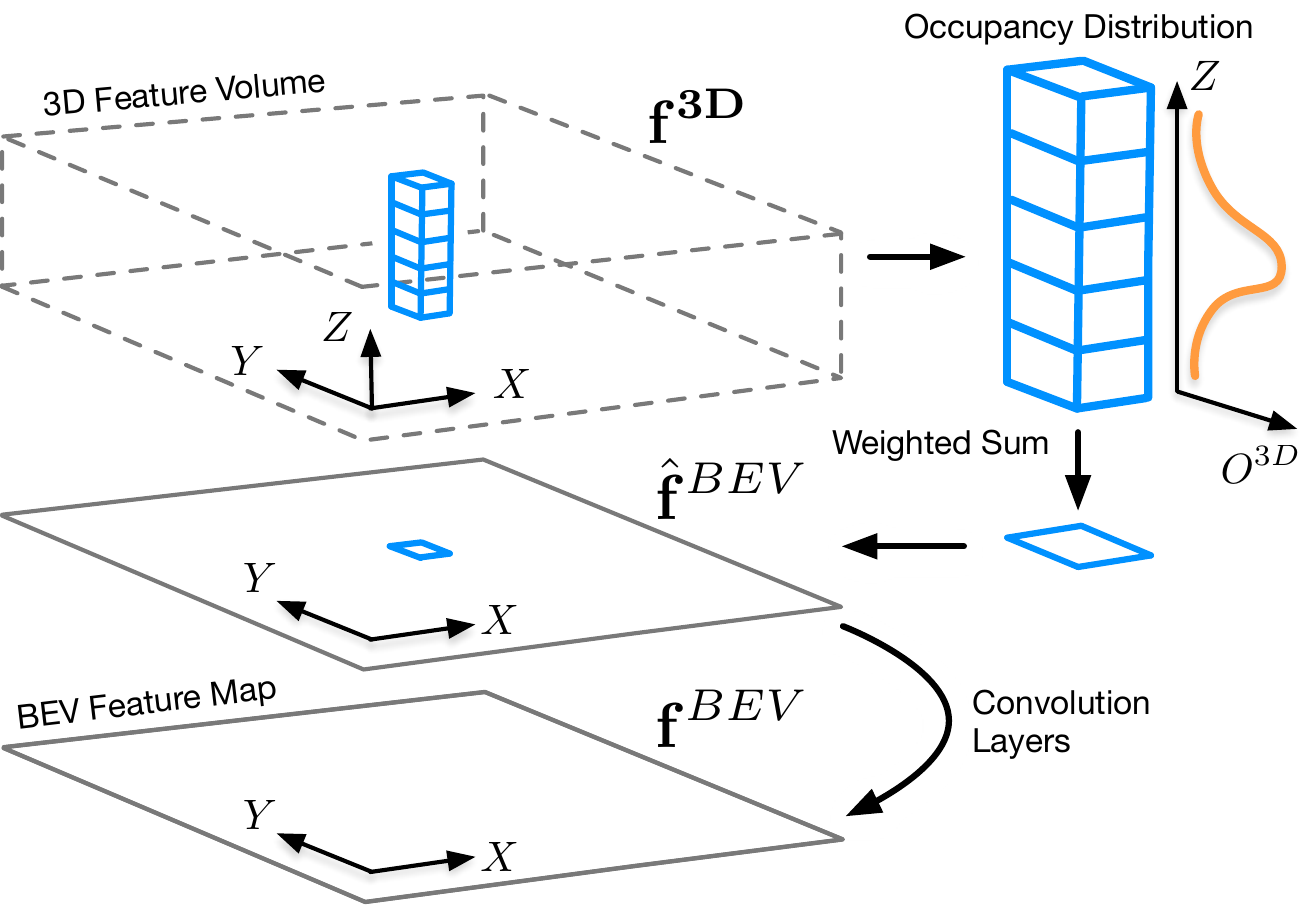}
    \vspace{-0.1cm}
	\caption{Occupancy aware feature aggregation}
	\label{fig:compression}
	\end{center}
	\vspace{-0.9cm}
\end{figure}

As shown in Fig.~\ref{fig:compression}, the 2D BEV coordinate system is aligned with the $XY$ plane of the ego vehicle coordinate system where the shared origin is defined on the center of the ego vehicle. Note that the BEV coordinate system only has 2 dimensions, forgoing the $Z$ dimension. The goal of the feature aggregation is to transform the 3D feature volume in ego vehicle coordinate into a 2D feature map in the BEV space, which can be treated as aggregating the 3D feature volume along its Z axis. To this end, we first rearrange the previously computed depth likelihood for all voxels by Eq.~\ref{eq:likelihood} into a depth likelihood volume $P^{3D} \in \mathbb{R}^{X'\times Y'\times Z'}$, which shares the same volumetric coordinate as that of 3D feature volume ${\bf f}^{3D}$. For each column along the Z-axis in the depth likelihood volume, the likelihood of each voxel of different height reflects its spatial occupancy. Thus, we normalize the depth likelihood along $Z$ axis into a spatial occupancy distribution, forming a spatial occupancy volume $O^{3D} \in \mathbb{R}^{X'\times Y'\times Z'}$ defined as
\begin{equation}\label{eq:occupancy}
    O^{3D}(x,y,z) =  \frac{P^{3D}(x,y,z) + b_o}{\sum_{z_i=0}^{Z'-1}P^{3D}(x,y,z_i) + b_o},
\end{equation}
where the $b_o$ is a bias term to encourage an equal contribution of feature on completely occluded region.

Our feature aggregation along the $Z$-axis could minimize the influence of features from empty voxels to the final feature in the BEV frame. Given the spatial occupancy volume $O^{3D}$, we compute the final 2D BEV feature as a weighted sum of 3D features
\begin{equation}\label{eq:compress}
    \hat{{\bf f}}^{BEV}(x,y) = \sum_{z_i=0}^{Z'-1} (O^{3D}(x,y,z_i)\times {\bf f}^{3D}(x,y,z_i)),
\end{equation}
\noindent where we use the normalized spatial occupancy distribution as the 3D feature weight.

We further transform $\hat{{\bf f}}^{BEV}$ via a few layers of convolution to obtain the final feature for BEV space ${\bf f}^{BEV}$ which is then applied to detection and segmentation tasks.

\subsection{Object Detection and Segmentation}
Given the BEV feature map, we use two heads for detection and segmentation. Specifically, we adopt the detection head and segmentation head from M$^2$BEV~\cite{m2bev} without modification for fair comparison. The detection head consists of three convolution layers and outputs dense 3D anchors in BEV space along with category, box size, and direction of each object. The segmentation head consists of five convolution layers and outputs 2 classes predictions, \textit{road} and \textit{lane}, as originally defined by LSS\cite{lss}.

\subsection{Training Strategy}\label{sec:sup}
We adopt supervised training strategy. We supervise the parametric depth estimation by maximizing its depth likelihood on ground-truth depth observations. Specifically, we minimize the negative log-likelihood loss $\mathcal{L}_{D}$ using sparse ground-truth depth $d_{gt}$ generated from sparse lidar measurements. Here $\mathcal{L}$ represent Laplacian distribution and $P^i_{gt}$ represent set of pixels where ground-truth lidar measurements is valid for image $i$.
\begin{equation}\label{eq:nll_loss}
    \mathcal{L}_{D}(\theta) =\sum_{i=1}^{N}\sum_{p\in \mathcal{P}^i}-\log(\mathcal{L}(d^p_{gt,i}|\mu_i^p(\theta), b_i^p(\theta)))
\end{equation}
where $\mathcal{P}^i$ defines the set of pixel coordinates with valid ground truth depth map for view $i$.

For detection head, we use the 3D detection loss used in PointPillars\cite{lang2019pointpillars} as follows, where $\mathcal{L}_{loc}$ is the total localization loss, $\mathcal{L}_{cls}$ is the object classification loss, $\mathcal{L}_{dir}$ is the direction classification loss, $N_{pos}$ refer to the number of positive samples and $\beta_{cls}, \beta_{loc}, \beta_{dir}$ are set to $1.0, 0.8, 0.8$ accordingly. 
\vspace{-0.1cm}
\begin{equation}\label{eq:det_loss}
    \mathcal{L}_{det} = \frac{1}{N_{pos}}(\beta_{cls}\mathcal{L}_{cls} + \beta_{loc}\mathcal{L}_{loc} + \beta_{dir}\mathcal{L}_{dir})
\end{equation}
Please refer to \cite{lang2019pointpillars} for more details.

For segmentation head, we use both Dice loss $\mathcal{L}_{dice}$ and binary cross entropy loss $\mathcal{L}_{bce}$ as segmentation loss $\mathcal{L}_{seg}$ and use equal weight $\beta_{dice} = \beta_{bce} = 1$.
\vspace{-0.1cm}
\begin{equation}\label{eq:seg_loss}
    \mathcal{L}_{seg} = \beta_{dice}\mathcal{L}_{dice} + \beta_{bce}\mathcal{L}_{bce}
\end{equation}

For the visibility map and additional outputs, since they are geometrically derived from the estimated parametric depth representation without any learned parameters, it's not necessary to apply supervision on them.

\section{Visibility}\label{sec:vis_eval}
\subsection{Visibility Map}
\label{sec:vis}

The segmentation in BEV space mainly focuses on segmenting lane regions. However, those regions are not always visible in the camera views due to the occlusion of vertical scene structures such as building (see Fig.\ref{fig:hallucination}). We thus propose to use our parametric depth modeling to infer a visibility map which decouples visible and occluded areas and, will contribute to mitigate the hallucination effect.

We define a visibility map $V^{BEV} \in \mathbb{R}^{X\times Y}$ to describe the visibility range of ego vehicle's multi-view cameras. Starting from the likelihood of the Laplacian distribution in Eq. \ref{eq:likelihood}, the occlusion probability $B(d)$ of a voxel in 3D space that has a back-projected depth $d$ in camera view is
\vspace{-0.2cm}
\begin{equation}\label{eq:vis1}
B(d) = \int_{0}^{d} \mathcal{L}(x|\mu,b) dx.
\end{equation}

\noindent We derive this occlusion probability as follows. Firstly we find the indefinite integral of Eq. \ref{eq:likelihood} as 
\begin{equation}\label{eq:vis2}
F(x) = \int_{-\infty}^{x} \mathcal{L}(x|\mu,b)dx = \begin{cases}
                        \frac{1}{2}\exp(\frac{x-\mu}{b}) \text{~if~} x < \mu\\
                        1-\frac{1}{2}\exp(-\frac{x-\mu}{b}) \text{~if~} x \geq \mu.
                    \end{cases}
\end{equation}

\noindent Then we calculate the definite integral between $[0,d]$ as the occlusion probability $B(d)$, which is defined as
$B(d) = F(d) - F(0) = F(d)-\frac{1}{2}\exp(-\frac{\mu}{b}).$

In practice, this is computed very efficiently, without the need to perform the discrete integration of the depth likelihood over the range $[0,d]$. Based on the relationship between visibility and occlusion, we convert the occlusion probability $B$ to visibility probability $V$ by 
\begin{equation}\label{eq:vis_prob}
    V(d) = 1-B(d) = 1 + \frac{1}{2}\exp(-\frac{\mu}{b})-F(d).
\end{equation}

To finally compute the visibility in BEV space, we take the maximum visibility probability along the $Z$ axis to form the visibility map $V^{BEV}$.

\begin{equation}\label{eq:vis_map}
    \tilde{V}^{BEV}(x,y) = \max_{z\in\mathcal{Z}'}V(x,y,z)
\end{equation}
where $\mathcal{Z}'=\{0,1,2\cdots Z'-1\}$. The ${V}^{BEV}$ is obtained via interpolation from $\tilde{V}^{BEV}$.
\subsection{Visibility-aware Evaluation}
For semantic segmentation where the ground-truth is usually generated using aerial images, it is not possible evaluate predictions in visible and occluded areas by using the standard evaluation metrics. Therefore, in this section, we follow a similar process as the one to generate the visibility map to derive a visibility-aware evaluation method for segmentation in BEV space. In this case, however, we project the lidar 3D points (ground-truth) into multi-view image space and use a depth completion network to obtain multi-view dense depth maps. This depth map is then used as the expected depth value to build a parametric depth representation $F(\theta_{gt})$. We then evaluate the ground-truth depth likelihood on each voxel in 3D space using Eq.~\ref{eq:likelihood}, forming the ground-truth depth likelihood volume $L_{gt}$. Finally, we derive the ground-truth visibility map in BEV space $V$ using Eq.~\ref{eq:vis_prob} and Eq.~\ref{eq:vis_map}.

In this case, $V$ reflects the maximum visibility of the multi-view cameras in BEV space. Thus, it can be used as a mask to explicitly evaluate results in BEV space subject to visibility. Specifically, we use a threshold $\tau_{vis}$ to split the predicted segmentation $s_{pred}$ and ground-truth segmentation label $s_{gt}$ into visible region $\{s^{vis}_{pred},s^{vis}_{gt}\}$ and occluded region \textbf{$\{s^{occ}_{pred},s^{occ}_{gt}\}$}. We can then compute the IoU for the visible ($IoU_{vis}$) and occluded ($IoU_{occ}$) regions separately as 
$s^{vis} = \sum_{x\in \mathcal{X},y\in\mathcal{Y} }s(x,y)\times\;\mathbbm{1}(V(x,y) \geq \tau _{vis}),
$

~$s^{occ} = \sum_{x \in \mathcal{X}, y\in\mathcal{Y}}s(x,y)\times\mathbbm{1}(V(x,y) < \tau _{occ})
$,
$IoU_{vis} = \frac{s^{vis}_{pred}\cap s^{vis}_{gt}}{s^{vis}_{pred}\cup s^{vis}_{gt}},\; IoU_{occ} = \frac{s^{occ}_{pred}\cap s^{occ}_{gt}}{s^{occ}_{pred}\cup s^{occ}_{gt}}
$ where $\mathcal{X}=\{0,1,\cdots,X-1\}$,  $\mathcal{Y}=\{0,1,\cdots,Y-1\}$, and $\mathbbm{1}(\cdot)$ is the indicator function.
We also report the occlusion rate on nuScenes as the percentage of visible or occluded segmentation labels over total number of segmentation labels. 


%% file: experiments.tex
\begin{table*}[!t]
\vspace{-0.3cm}
  \centering
  \resizebox{0.7\textwidth}{!}{
  \begin{tabular}{l|cc|cccccc}
    \toprule
    \textbf{Camera-based} Methods  & mAP↑ & NDS↑ & mATE ↓ &	mASE ↓ & mAOE ↓	& mAVE ↓ & mAAE↓   \\
    \midrule
    CenterNet\cite{duan2019centernet} & 0.306 & 0.328 &0.716	&0.264	&0.609	&1.426	&0.658  \\
    FCOS3D\cite{wang2021fcos3d} & 0.343 & 0.415 &0.725	&0.263	&0.422	&1.292	&0.153  \\
    DETR3D\cite{wang2022detr3d} & 0.349 & 0.434 &0.716	&0.268	&0.379	&0.842	&0.200  \\
    PGD\cite{wang2022probabilistic} & 0.369 & 0.428 &0.683	&\textbf{0.26}	&0.439	&1.268	&\textbf{0.185}  \\
    M$^2$BEV\cite{m2bev} &  0.417 & 0.470 &0.647	&0.275	&0.377	&0.834	&0.245  \\\hline
    BEVFormer\cite{li2022bevformer} (single-frame version) &  0.417  & 0.448 &0.647	&0.275	&0.377	&0.834	&0.245  \\
    BEVFusion\cite{liu2022bevfusion} (camera-only version) &  0.417 & 0.321 &0.647	&0.275	&0.377	&0.834	&0.245  \\\hline
    Ours  & \textbf{0.436} & \textbf{0.496} &\textbf{0.637}	&0.264	&\textbf{0.367}	&\textbf{0.810}	&0.194  \\
    \bottomrule
  \end{tabular}
  }
  \vspace{-0.2cm}
  \caption{\textbf{Detection results on the nuScenes validation set.} We report our results compared to other camera-based methods. Our approach outperforms existing methods for all the metrics except for mASE and mAAE where the performance is slightly lower than PGD~\cite{wang2022probabilistic}.}
  \label{table:det_val}
  \vspace{-0.4cm}
\end{table*}

\begin{table}[!t]
  \centering
\resizebox{0.8\columnwidth}{!}{
  \begin{tabular}{cl|c|cc}
    \cmidrule[\heavyrulewidth]{2-5}
    & Methods  & Modality & mAP↑ & NDS↑ \\
    \cmidrule{2-5}
    & PointPillars\cite{lang2019pointpillars} &  Lidar & 0.305 & 0.453  \\
    & CenterFusion\cite{nabati2021centerfusion} &  Camera + Lidar & 0.326 & 0.449  \\
    & CenterPoint v2\cite{yin2021center} &  Camera + Lidar + Radar & 0.671 & 0.714  \\
    \cmidrule{2-5}
     & CenterNet\cite{duan2019centernet} &  Camera & 0.338 & 0.400  \\
      & FCOS3D\cite{wang2021fcos3d} & Camera & 0.358 & 0.428  \\
     & DETR3D\cite{wang2022detr3d} & Camera & 0.349 & 0.434  \\
    & PGD\cite{wang2022probabilistic} & Camera & 0.386 & 0.448  \\
    & M$^2$BEV\cite{m2bev} &  Camera & 0.429 & 0.474  \\
    & Ours &  Camera & \textbf{0.468} & \textbf{0.495}  \\
    \cmidrule[\heavyrulewidth]{2-5}
  \end{tabular}
  }
  \vspace{-0.1cm}
  \caption{\textbf{Detection results on nuScenes test set.} Our method outperforms existing camera based methods for both mAP and NDS.}
  \label{table:det_test}
  \vspace{-0.1cm}
\end{table}

\begin{table}[!t]
  \centering
\resizebox{0.6\columnwidth}{!}{
  \begin{tabular}{l|ccc}
    \toprule
    \textbf{Camera-based} Methods  & Road ↑ & Lane ↑ \\
    \midrule
    CNN\cite{lss}  & 68.9 & 16.5  \\
    Frozen Encoder\cite{lss} & 61.6 & 16.9  \\
    PON\cite{pon} & 60.4 & -  \\
    OFT\cite{roddick2018orthographic} &  71.6 & 18.0  \\
    LSS\cite{lss} & 72.9 & 19.9  \\
    M$^2$BEV\cite{m2bev} & 77.2 & 40.5  \\
    Ours* (w/o depth sup.) & 77.9	& 40.8  \\
    Ours & \textbf{78.7}	& \textbf{41.0}  \\
    \bottomrule
  \end{tabular}
  }
  \caption{\textbf{Segmentation results on the nuScenes validation set.} We report our results compared to other camera-based methods. Our approach outperforms all existing approaches in the literature including M$^2$BEV, demonstrating the benefit of our feature transformation module.}
  \label{table:nuscene_seg}
  \vspace{-0.3cm}
\end{table}

\begin{table}[!t]
  \centering
  \resizebox{\columnwidth}{!}{
  \begin{tabular}{l|cc|cc|cc}
    \toprule
       & \multicolumn{2}{|c|}{Vis. (66.9\%)}  & \multicolumn{2}{|c|}{Occ. (33.1\%)} & \multicolumn{2}{|c}{All region} \\
    Methods & Road↑ & Lane↑ & Road↑ & Lane↑ & Road↑ & Lane↑ \\
    \midrule
    LSS\cite{lss}  & 79.4 & 23.1 & 47.1 & 6.5 & 72.9 & 19.9  \\
    M$^2$BEV\cite{m2bev}  & 82.9	&39.8	&48.9	&12.8	&73.2	&36.1 \\
    Ours  & \textbf{84.8}	&\textbf{41.9}	&48.9	&12.4	&\textbf{73.8}	&\textbf{36.5}  \\
    \bottomrule
  \end{tabular}
  }
  \vspace{-0.1cm}
  \caption{\textbf{Segmentation results on the nuScenes validation set under visibility constraints.} We decouple the evaluation of the segmentation results on NuScenes validation set into visible areas (66.9\%) and occluded areas (33.1\%) based on the visibility map. Our approach performs on par on hallucinated areas and, importantly, for visible areas yields significant improvements compared to existing methods. }
  \label{table:seg_vis}
  \vspace{-0.1cm}
\end{table}

\begin{table}[!t]
  \centering
  \resizebox{0.8\columnwidth}{!}{
  \begin{tabular}{l|cccc}
    \toprule
    Methods & mAP↑ & NDS↑ & Road↑ & Lane↑\\
    \midrule
    M$^2$BEV\cite{m2bev} & 0.408 & 0.454 & 75.9 & 38.0 \\
    Ours & \textbf{0.424} &	\textbf{0.467} &	\textbf{76.5} &	\textbf{39.9} \\
    \bottomrule
  \end{tabular}
  }
  \vspace{-0.1cm}
  \caption{\textbf{Joint detection and segmentation results on the nuScenes validation set.} We report joint estimation results for segmentation and detection and compare our results to M$^2$BEV. Our method outperforms the baseline for all the metrics. }
  \label{table:nuscene}
  \vspace{-0.1cm}
\end{table}
\begin{table}[!t]
  \vspace{-0.2cm}
  \centering
    \resizebox{0.8\columnwidth}{!}{
  \begin{tabular}{l|cccc}
    \toprule
    Model & mAP↑ & mIoU↑ & FPS & Memory \\
    \hline
    M$^2$BEV\cite{m2bev} & 0.408 & 56.9 &1.2 & 7718  \\
    Ours & \textbf{0.424} & \textbf{58.2} & 1.3 & 8902  \\
    \bottomrule
  \end{tabular}
  }
  \vspace{-0.2cm}
  \caption{\textbf{Performance analysis}. We report frames per second (FPS) and memory requirements for our model and M$^2$BEV when running on a Nvidia Titan V100 GPU.}
  \label{tab:efficiency}
  \vspace{-0.5cm}
\end{table}

\section{Experiments}
In this section, we first detail our experimental settings, then we demonstrate the effectiveness of our approach on the nuScenes dataset, and, finally, we provide ablation studies on the main components of our method. 
\subsection{Implementation Details}
\noindent\textbf{Dataset.} We conduct our experiments on the nuScenes dataset~\cite{NuScenes}. The nuScenes dataset provides video sequences along with multiple sensor outputs including Lidar, Radar, GPS and IMU, all of which are collected by calibrated and synchronized sensors mounted on an vehicle driving across Boston and Singapore. The dataset consists of 1000 sequences, split into 700 for training and 150 for validation and testing, respectively. Each sample provides six RGB images captured by 6 cameras with divergent viewing directions along with Lidar sparse 3D points, Radar sparse 3D points, GPS pose and IMU readouts. We follow~\cite{lss,m2bev} to generate ground-truth segmentation labels from the global map provided by nuScenes dataset.

\noindent\textbf{Evaluation metrics.} We report our results using the same metrics as in the nuScenes benchmark. For detection, we report mean Average Precision (mAP) and the nuScenes detection score~\cite{NuScenes}. For segmentation, we follow LSS~\cite{lss}, and report the mean IoU score (mIoU). In addition, we report results using the proposed visibility-aware evaluation detailed in Sec.~\ref{sec:vis_eval}. Unless specified, we report numbers on the validation set. 

\noindent\textbf{Network architecture.} We use a unified framework to demonstrate benefits of our depth-based feature transformation module. The network consists of a backbone image encoder and two decoding heads, one for segmentation and one for detection. We use ResNet with deformable convolution as the image encoder. For the decoding heads, we use the same architecture as the one in PointPillars~\cite{lang2019pointpillars}. 

We set the size of the intermediate 3D volume consisting of $X'\times Y'\times Z' = 400\times400\times12$ voxels, with a voxel size of $0.25m\times 0.25m\times 0.5m$, respectively. The final BEV space dimension consists of $X\times Y = 200\times200$ grids. Each grid is of size $0.5m\times 0.5m$.

\noindent\textbf{Training and inference.} During training, we use 6 RGB images and corresponding camera parameters as input. 
The training for parametric depth estimation is supervised by the ground-truth sparse Lidar points provided in the dataset. Ground-truth detection and segmentation labels are used to supervise the detection and segmentation heads. We set batch size to 1 per GPU and use 3 nodes with 8 Nvidia V100 GPUs. For inference, our method only requires the 6 input RGB images together with the corresponding camera parameters. 

\subsection{Results}
We now compare our results with M$^2$BEV and other state-of-art methods on the nuScenes dataset. To facilitate the comparison to other approaches, we use ResNeXt-101 as the backbone of our method for detection and segmentation experiments and use ResNet-50 as the backbone for multi-task learning experiments and efficiency analysis.

\noindent\textbf{Detection.} We report the results of our method and related state of the art methods in Tab.~\ref{table:det_val} and Tab.~\ref{table:det_test}, for the validation set and the test set respectively. For the validation set, we only include frame-wise camera-based methods. That is, we exclude those approaches using temporal information. For the test set, we include the latest results including Camera, Lidar, Radar and their combination. As we can see, in both sets, our approach outperforms all existing camera-based methods on both mAP and the NDS score.

\begin{table}[!t]
  \centering
  \vspace{-0.1cm}
    \resizebox{0.85\columnwidth}{!}{
  \begin{tabular}{llcccc}
    \toprule
    Lift & Aggregate & mAP & NDS & Road & Lane \\
    \midrule
    PON\cite{pon} & PON\cite{pon} & - & - & 60.4 & -  \\\hline
    Non-parametric\cite{lss} & PP\cite{lang2019pointpillars} & 0.409 & 0.455 & 75.9 & 37.9 \\
    Non-parametric\cite{lss} & Our Occupancy & \underline{0.414} & \underline{0.459} & \underline{76.1} & \underline{38.3} \\\hline
    Uniform\cite{m2bev} & PP\cite{lang2019pointpillars} & 0.408 & 0.454 & 75.9 & 38.0  \\
    Uniform\cite{m2bev} & Our Occupancy & 0.413 & \underline{0.459} & 76.0 & 38.2  \\\hline
    Our Parametric depth & PP\cite{lang2019pointpillars} & 0.410 & 0.457 & 76.0 & 38.0  \\
    Our Parametric depth &  Our Occupancy & \textbf{0.424} & \textbf{0.467} & \textbf{76.5} & \textbf{39.9}  \\
    \bottomrule
  \end{tabular}
  }
  \vspace{-0.1cm}
  \caption{\textbf{Ablation Study.} Influence of the different components of our feature transformation approach and their comparison to other methods available in the literature.}
  \label{table:ablation_feat_trans}
  \vspace{-0.4cm}
\end{table}

\noindent\textbf{Segmentation.} We now focus on evaluating our semantic segmentation results. We report our performance compared to state-of-the-art methods on the nuScenes validation set in Tab.~\ref{table:nuscene_seg}. 
We also report a variant of our model trained without depth supervision (Ours*) to fairly compare with LSS~\cite{lss}.
Our method performs significantly better compared to LSS~\cite{lss} on both road and lane segmentation and slightly better compared to M$^2$BEV~\cite{m2bev}, the closest method to ours. 
Our model without depth supervision still outperforms existing methods.
Interestingly, if we take the visibility into account, as shown in Tab.~\ref{table:seg_vis} and Fig. \ref{fig:hallucination}, our method clearly outperforms the baselines on the visible areas while maintain the performance compared to M$^2$BEV on the occluded regions. These results evidence the benefits of our parametric depth approach.

\noindent\textbf{Joint detection and segmentation.} Finally, we report results for jointly evaluating both tasks. In this case, we compare our results to the multi-task version of M$^2$BEV. We show results for this experiment in Tab.~\ref{table:nuscene}. Our method, once again, outperforms the baseline on both detection and segmentation tasks. These results further evidence the benefits of an improved depth representation in the 2D to 3D feature transformation process. 

\noindent\textbf{Efficiency.} Our parametric depth estimation requires the estimation of additional parameters compared to simplified depth estimation approaches. As shown in Tab.~\ref{tab:efficiency}, our model requires slightly larger amount of memory; However, that does not lead to a significant increase in the inference time.

\subsection{Ablation Studies}
We carry out ablation experiments to study the influence of feature transformations on final detection and segmentation performance and the robustness of our model to calibration error. More ablation experiments can be found in supplementary material. We use ResNet-50 as the backbone for all ablation experiments.

\noindent\textbf{Feature transformations}
We evaluate the effectiveness of the parametric depth based feature lifting and aggregation module comparing with baseline non-parametric depth based lifting LSS\cite{lss}, baseline uniform depth based lifting similar to M$^2$BEV and the widely used Pointpillar\cite{lang2019pointpillars} feature aggregation. Results are in Tab. \ref{table:ablation_feat_trans}. Our proposed parametric depth based lifting coupled with occupancy based feature aggregation achieved best performance for both detection and segmentation.

\noindent \textbf{Limitations.} Like all camera based methods, our method can only provide reliable detection and segmentation results on visible region. On occluded region, although our method can provide hallucination results and visibility information, the results are not reliable for making critical driving decision. Following planning tasks should utilize the visibility and uncertainty information to achieve reliable planning.


%% file: conclusion.tex
\vspace{-0.20cm}
\section{Conclusion}
\vspace{-0.1cm}

 We propose a parametric depth distribution modeling-based feature transformation that efficiently transforms 2D image features to BEV space. By incorporating visibility inference, our method can provide crucial visibility information to down-streaming planning tasks. Moreover, our approach outperforms existing methods in both detection and segmentation tasks, making it a promising candidate for feature transformation in future works. In our future work, we aim to investigate the integration of temporal information to improve estimation accuracy.